# A Fusion Model: Towards a Virtual, Physical and Cognitive Integration and its Principles

Hao Lan Zhang Member, IEEE, Yun Xue, Yifan Lu, and Sanghyuk Lee, Senior Member, IEEE

*Abstract*— Virtual Reality (VR), Augmented Reality (AR), Mixed Reality (MR), digital twin, Metaverse and other related digital technologies have attracted much attention in recent years. These new emerging technologies are changing the world significantly. This research introduces a fusion model, i.e. Fusion Universe (FU), where the virtual, physical, and cognitive worlds are merged together. Therefore, it is crucial to establish a set of principles for the fusion model that is compatible with our physical universe laws and principles. This paper investigates several aspects that could affect immersive and interactive experience; and proposes the fundamental principles for Fusion Universe that can integrate physical and virtual world seamlessly.

*Index Terms*—Virtual Reality (VR), Fusion, Integration, Virtual World, Human-computer Interaction

## I. Introduction

The rapid development of virtual technologies draws a blueprint of our future life. The extensive applications of VR/AR/MR accelerate the integration process of virtual worlds with our physical world. These applications are particularly successful in several domains including tourism, building information modeling, automotive, training and education, healthcare, and urban infrastructure [1, 2]. During the COVID-19 period, the demands for integrating virtual technologies with our physical world applications are rising quickly due to the physical contact become difficult for many people. The combination of online and offline model is becoming popular daily norm for many people for their work, study, business and entertainment. Some researchers consider the new form of integrating online and offline working style as the future trends of our society, which inspired the development of the new fusion model, i.e. Fusion Universe.

The major difference between Metaverse and Fusion Universe is that Metaverse tends to be more virtual and digital oriented, whereas Fusion Universe is more physical-extension oriented. In other words, Metaverse and other virtual reality applications aim to digitally replicate nature environment for immersive user experience, which are considered as derivative of the real world. Fusion Universe, on the other hand, extends our physical world into a physically, cognitively and digitally merged world, where our physical universe is extended. Therefore, FU contains our physical world and its seamlessly integrated virtual environments. The following figure shows the relationship of Fusion Universe, virtual world and physical world.

The most significant advantage of using MRI for brain analysis is that MRI can produce high spatial resolution images, which allow brain functions to be associated and analyzed spatially. Fig 1 illustrates the example of high spatial resolution brain images using MRI [1, 2]. However, MRI has several critical disadvantages including low temporal resolution, magnetic field risk (people with implants), high costs, etc. These limitations hindered the development of MRI technology for brain analysis. Whereas, EEG can overcome all the above disadvantages of MRI. Therefore, many researchers are keen to utilize EEG as one of the major technologies for brain analysis, and find the solution for improving the spatial resolution of EEG.

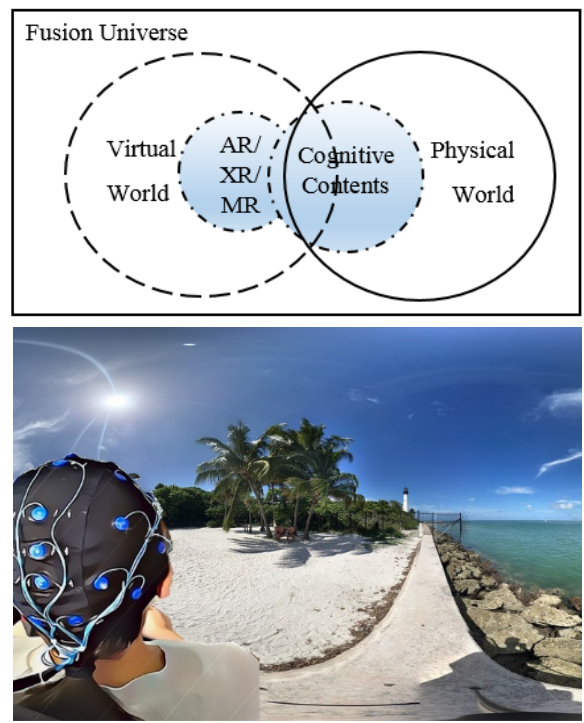

Fig 1 The relationship of Fusion Universe, virtual world, physical world and cognitive contents.

Cognitive information is regarded as a part of physical content and a part of informative content. Most of cognitive contents are not materialized such as skills, feelings, instinct, intellectual properties, etc. These informative contents are

Manuscript received. This work is partially supported by Zhejiang Natural Science Fund (LY19F030010), Zhejiang Philosophy and Social Sciences Fund (20NDJC216YB)

H.L. Zhang is with the Center for SCDM, NIT, Zhejiang University, China (e-mail: haolan.zhang@nit.zju.edu.cn).

intangible until they are converted to tangible forms include texts, drawings, cognitive controlled systems, etc. However, cognitive contents are not virtual although they can be delusional sometimes. In this paper, cognitive information is regarded as the combination of virtual and physical contents. Cognitive contents naturally fuse information generated in physical and virtual worlds. The recent proposed Cognitive Digital Twin (CDT) concept is a typical integration of digital/virtual and physical applications [3].

Recent research work reveal that current VR/AR/MR technologies are facing some challenges and issues, which include cyber-sickness, low user acceptance, user theme discrepancies, cyber security issue, head-mounted VR displayer resolution problem, ethic issues, etc. [1-2, 4]. Much work has been done to investigate user feedbacks and experiences of using virtual reality applications [4, 5]. These investigations indicate that the current VR/AR/MR technologies are somehow primitive and flawed. The related virtual technologies are not only primitive in hardware but also in the overall theoretical basis. Therefore, it is crucial to understand the challenges in virtual worlds and provide systematic guidelines and principles for integrating virtual and physical worlds, i.e. Fusion Universe.

## II. Discrepancies in Virtual Worlds

Nowadays, virtual and immersive technologies have been applied in many areas of our physical world. However, some researchers found that current VR/MR devices may cause health problems for some users including dizzy, visual fatigue, hearing loss, etc. The health issues represent the typical problem of integrating virtual world with physical world. The other notable problems include the lack of consistency with physical world, the lack of standardize regulation, etc. Some of these problems are partially caused by low quality of hardware, however, we believe the discrepancy between virtual world and physical world is one of the major obstacles that hinders the development of virtual world technologies.

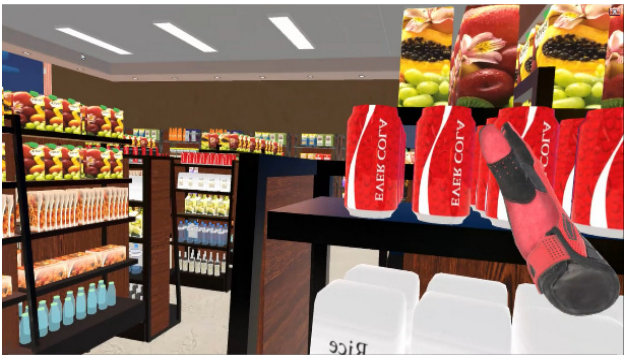

Figure 2. A VR shop demo is created for evaluation and observation in this research.

A VR shopping scenario has been developed in this research to understand virtual life experience as shown in Figure 2. Furthermore, an immersive VR scene has been developed with real photos rendered in the scene as shown in Figure 3. The preliminary evaluation conducted in this paper indicates that users can obviously perceive the inconsistency in virtual world. For instance, user actions cannot seamlessly interact with its environment as shown in Figures 3. The objects in VR environment do not follow the law of physics in our real life [6]. In Fusion universe, the discrepancy between physical and virtual worlds should not be perceived by users.

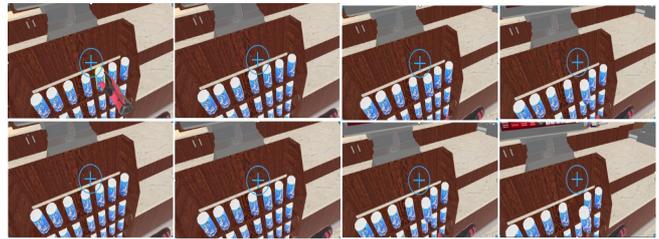

Fig. 4: An observation on the VR shopping scenarios developed in this research. The observation is conducted in frame by frame basis; and discrepancy has been detected when the VR hand tries to grab a product from the shelf. The product moves without hand holding.

We further evaluated the interaction in AR environment. The similar discrepancy can be detected in AR environment as shown in Enriched city AR scenario [7]. The visual effects on physical products are not coherent. The similar findings have been observed in [8, 9]. Fusion Universe will provide users a unified platform, where physical space and virtual illusion can merge seamlessly. In the foreseeable future, physical-digital-integration platforms, such as Fusion universe, will play a major role in our daily life. In order to create an integrated environment in FU, a set of principles are defined in the next section.

## III. The Principles of Fusion Integration

Our world obeys the laws of nature, which include the Newton's laws of motion, the big bang theory, Archimedes' principle, Darwin's evolution theory, etc. In order to create a seamlessly merging environment in Fusion universe, which allows users to switch between physical and virtual worlds without feeling inconsistency. To achieve this goal, the following aspects need to be addressed in Fusion universe.

### A. *The principle of limits in the fusion process:*

In the real world, people's life is constrained by various limitations. For instance, under normal light conditions, human retina requires approximately 80 milliseconds for recording a new vision information [10]; the maximum speed limit in the universe is 299,792.458 kilometers per second; the smallest particle in the world is quarks, which is $10^{-19}$ meters and top-quark life time is $\approx 0.5 \times 10^{-24}$. In Fusion universe, various limits should be defined in regard to size, speed, latency, distance, etc.

The calculation of spatial distance in Fusion Universe should be unified in both physical and virtual environments in order to deliver a seamless fusion process. In virtual scenarios, distances between objects are normally based on the pixel size of VR displayers. In digital worlds, space zooming would alter the pixel quantity, which might cause discrepancy in distance calculation or interaction with the physical world. Hence, the limits laws are required in virtual world. The distance in Fusion Universe is based on spatial mapping/meshing as shown in Figure 5. The smallest element for distance calculation is pixels, where pixel size $p \rightarrow 10^{-19}$ (pixel size approaches to quark's size). Therefore, the following size of FU can be defined in (1), which means FU is an infinite space.

$$\lim_{n\to\infty} f(p_n) = \lim_{n\to\infty} \sum_{i=1}^{n} p_i = \infty \qquad (1)$$

where *f(p$_n$)* is linear accumulation of smallest particles in FU. The number of the particle approaches infinite, which aligns with the physical world; and allows virtual and physical spaces to be fused smoothly.

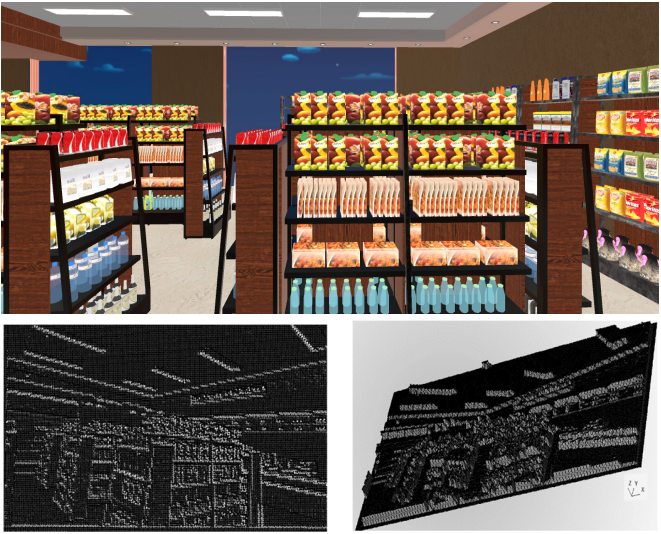

Figure 5. VR Spatial Mapping. The figure on top is our VR shop and the bottom figures are the VR shop's spatial mapping/meshing view by rotating the axes.

Time can be fragmented in Fusion Universe to synchronize virtual, cognitive and physical worlds. Suppose time fragmentations approach to 0 in FU, an object is throwing from the physical world to a virtual world in Fusion Universe as shown in Figure 6. The total time of the object traveling in air before it hits ground should be a constant value. Therefore, it can be calculated as Equation 2. Although, some would argue that the most efficient way is to set the time in physical world as the reference for FU. However, if virtual world and FU do not have its own time coordinate, the discrepancy will happen.

$$\lim_{n\to\infty} f(t_n) = \lim_{n\to\infty} \sum_{i=1}^{n} t_i = T \qquad (2)$$

where *f(t$_n$)* is linear accumulation of time fragmentation in FU. The number of the time fragment approaches infinite, which means each time fragment approaches to 0. Equation 2 describes the calculation of a particular event, nevertheless, time to future should be infinite in FU.

*B. The principle of probability in the fusion process:*

Probability theory plays a key role in our daily life, which defines the likelihood of occurrence of an event [11]. Probability is a central concept in scientific models of causation, which frequently change people's daily life. For instance, the probability of a person struck by lightning is very small. However, a tourist accidentally walking in a prairie in a thunderstorm weather will have a high probability of being struck by lightning. A lot of small things will contribute to the result of a tourist being struck by lightning, which is called butterfly probability [12]. If the tourist changes the travel time or miss the buses to the prairie, then the probability of being struck will be very low.

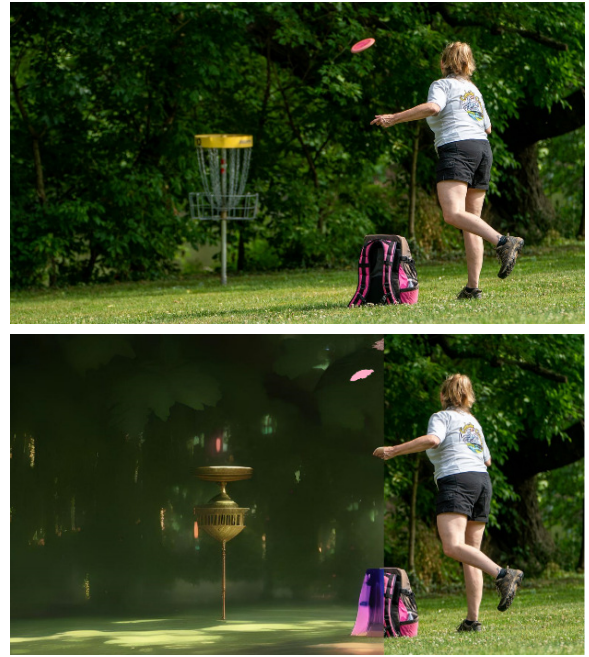

Figure 6. Throwing objects in Fusion Universe between virtual and physical worlds. The photo on the left is taken in the physical world and the right picture indicates the fusion scenario integrating virtual and physical worlds (the top photo is free to use under the Content License from pixabay.com).

In Fusion Universe, defining a comprehensive probability principle is crucial to fuse virtual and physical worlds seamlessly in FU. For independent events in FU, the probability in virtual/physical world and FU is statistically independent. In FU, the set of all possible outcomes denoted by *Ω*. Suppose *x* is an element of *Ω*, the probability function *f(x)* satisfies the following properties:

$$f(x) \in [0, 1] \text{ for all } x \in \Omega;\ \sum_{x \in \Omega} f(x) = 1. \qquad (3)$$

where the sum of probability function *f(x)* over all element x in *Ω* is equal to 1. *Ω* denotes the sample space with the set of all possible outcomes.

The probability distribution in FU can have some extreme situations. For example, if you toss a coin 10,000 times in FU, in extreme cases, the physical world can have 5,000 heads, and the virtual world can have 5,000 tails. But for FU it is still a fair distribution as a whole. However, most of events in our physical world are complicated and affected by various factors, which currently can be fused by Bayesian theorem (naïve classifier) as the follow.

$$P(E|F_1, \ldots F_n) = \frac{P(F_1, \ldots F_n|E)P(E)}{P(F_1, \ldots F_n)},$$
$$P(F_1, \ldots F_n) \neq 0. \qquad (4)$$

where all features $P(F_1, \ldots F_n)$ are mutually independent, conditional on event E.

The butterfly probability concept is introduced when several events with extremely small probabilities occurring randomly will eventually have big impact on results. The definition of Butterfly Probability (BP) is [13]: the BP of r occurring within time *t* is the value that $P_\varepsilon(r, U_0, t)$ converges to as $\varepsilon \to 0$, before it collapses to 0 or 1. This BP concept is particularly useful for cognitive analytics and integration in FU. The reason is that the entropy of cognitive contents is arguably higher than physical contents [14]. This results in persistent changes in cognitive systems, which will accumulate a significant impact in FU.

Furthermore, the Probabilistic Data Association (JPDA) can be used for instant object tracking and heterogeneous data fusion in FU, which is a suboptimal approach to the Bayesian filter [15]. Another useful tool, Permutation Entropy (PE), which is based on the probability theory to deal with time series data sources. PE is efficient for cognitive fusion, which can capture dynamics in time series data source, such as Electroencephalogram (EEG) based brain signals.

*C. Other fusion principles:*

Due to the page limitation, this paper only outlines several principles in FU. Some other factors need to be considered are:

(1) The FU model is centerless. Various newly developed virtual applications can integrate to FU; and form an Internet-like centerless FU.

(2) The FU model is superluminal. Objects in FU can travel fast than light through deploying portal-like transportation method, which allows objects instantly travel to another virtual location in FU.

(3) FU allows randomization regulated by probability. Human digital identities can mutate in FU. Each person entering a virtual environment from the physical world will be assigned a unique digital identity which can have slight mutation when virtual environment is changing.

## IV. Implementing the Principles

This section will illustrate how the fusion principles impact the digital models as a part of FU from users' perspective.

**Scenario 1**. This paper defines FU has the same object life-cycle as physical world. A new user will receive a unique digital identify before entering to FU, which is regarded as the birth process in FU. The creation of a digital human has a probability of random mutation in each part of the body, which will generate a unique digital human. The FU probability principle is crucial for uniquely creating a digital human in FU, which is similar to the human DNA mutation process.

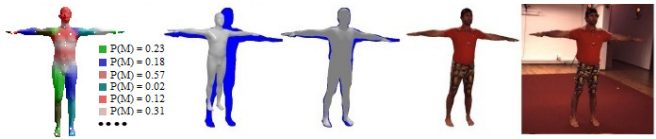

Figure 7. In this research, a unique digital identification is assigned to a digital human based on the random mutation regulated by the probability principle. A total of 24 segments (3D mesh) are generated for the digital human construction (shown on the first left digital human in color).

In Figure 7, the digital human is comprised of 24 segments [16]. Each segment highlighted with different colors and assigned with a probability value of mutation, e.g. P(M) = 0.23, which indicates the probability of slightly change to its assigned hash value. The final digital human has a unique identity that is comprised of 24 independent values. The 3D mesh of each segment is converted to a hash value, and fuse 24 values ($H_1 \otimes H_2 \otimes \ldots \otimes H_{24}$) to generate digital human identifies. In addition, the number of segments can be adjusted depending on user requirements.

Furthermore, the covariance matrix adaptation annealing method deployed in this digital human reconstruction process utilizes error approximation. Replicating 3D mesh sampling until samples converge to an optimum through adjusting resampling weights. The weight is formulated in a modified form of the Boltzmann distribution $\omega = \exp(-\lambda E(x,y))$ with respect to an error metric of $E(y, x)$. $\lambda$ is an annealing variable. Thus, when $\lambda \to \infty$, the mass of all weights will concentrate on the minimum of $E(y, x)$[16]. The process demonstrates the application of the limit principle in the digital human reconstruction.

**Scenario 2**. Non-invasive brain signals, i.e. EEG, can be used to reflect users' cognitive statuses in FU cognitive fusion process. The cognitive psychology data collected by EEG devices is to be analyzed and delivered to virtual and physical worlds. One of the efficient methods for EEG analytics is Permutation Entropy (PE), which provides time series data discrimination through calculation of the signal entropy based on probability theory.

A set of time series data $C = \{c_1, c_2, \ldots, c_n\}$, the conventional patterns of permutation have the following structure $P = \{p_1, p_2, \ldots, p_{n!}\}$. For the particular pattern $\pi_t$, where $\pi_t = (c_t, c_{t+\tau}, \ldots, c_{t+(n-1)\tau})$, $n$ is order from the time series $C$ and $\tau$ denotes distance between samples [17]. The PE of order $2 \leq n$ can be defined as:

$$H(n) = -\sum p(\pi)\log(p(\pi)) \quad (5)$$

where *n* is the order of the PE. The occurrences of the order pattern $\pi_i$ is denoted as $S(\pi_i)$, $i$= 1, 2, ... n!, $p(\pi)$ is the relative frequency which can be expressed as:

$$p(\pi) = \frac{C(\pi)}{T-(n+1)\tau} \quad (6)$$

The utilization of PE in the fusion model can efficiently analyze the complexity of cognitive time series data. It can be used to identify the status of a user's brain activity in virtual or physical worlds; and provide a low computation cost solution to the cognitive integration in FU. Figure 8 illustrates a fused system integrating the EEG analytics and VR application.

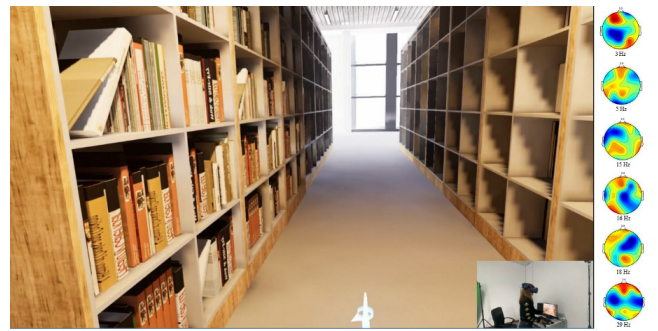

Figure 8. Permutation Entropy for EEG complexity calculation while user using the VR equipment. The EEG topographic maps on the left indicates brain complexity statuses at different frequencies.

## V. Conclusion and Future Perspective

Practically, virtual and cognitive contents are associated with our physical world, however, they do not necessarily exist in the physical world. This may result in discrepancy among physical, virtual and cognitive worlds. Therefore, the fusion model (FU) and several related principles are introduced in this paper to create an immersive environment for seamlessly integration and fusion of all three spheres. The concept of FU elevates our physical world to a higher dimension through merging virtual and cognitive contents into the physical world, which will extend and accelerate the development of our current world.

This paper describes several principles in FU. However, only several principles are outlined due to the page length limit. The future work will further extend the current work to develop comprehensive principles for FU.

ACKNOWLEDGEMENT

This work is partially supported by Ningbo Natural Science Fund Plan (Grant No. 2022J164), the Humanity and Social Science Foundation of the Ministry of Education of China (21A13022003), Zhejiang Provincial Natural Science Fund )LY19F030010(, the Ningbo Public Welfare Science and Technology Plan (Grant No. 2021S093).